\documentclass[11pt,a4paper]{article}

\usepackage[margin=1in]{geometry}
\usepackage{graphicx}
\usepackage{booktabs}
\usepackage{amsmath,amssymb}
\usepackage{natbib}
\usepackage{hyperref}
\usepackage{xcolor}
\usepackage[expansion=false]{microtype}
\usepackage{setspace}
\usepackage{enumitem}
\usepackage{xurl}
\usepackage[htt]{hyphenat}  

\hypersetup{
  colorlinks=true,
  linkcolor=blue!50!black,
  citecolor=blue!50!black,
  urlcolor=blue!60!black
}

\title{How to Build a Benchmark That Actually Measures Taste}

\author{%
Xiaofan Wu\thanks{Corresponding author: \texttt{xiaofan@chance.vision}}, Xi Zeng, Miaoxia Chen, Peishan Chen,\\
Shuyan Li, Chen Wang, Xiaohan Wang, Wanling Xu,\\
Jiyun Yao, Hanyong Zhong, Jiahao Zhu\\[4pt]
Chance AI
}

\date{}

\begin{document}
\maketitle

\begin{abstract}
Expert-aligned ``taste'' benchmarks should reward craft judgment, not the ability to game the test. We introduce \textbf{TasteBench}, a pairwise good-vs-bad benchmark across \textbf{visual}, \textbf{brand}, \textbf{book}, and \textbf{tech} items under a frozen rubric with expert-only gold (panel $n \ge 3$; \textbf{percent agreement $\ge 0.60$} primary gate; weak engagement labels train/analysis-only), together with a \textbf{control protocol} that separates craft judgment from test-gaming. Applying the protocol to a Round-1 pilot (\texttt{primary\_human\_expert}; three biological human experts; $n=250$ pairs) surfaces and quantifies \textbf{three artifacts} that undermine the benchmark's validity: \textbf{(1) position}---the preferred item is slot A with probability $0.956$, so an \texttt{always\_item\_a} policy scores $0.956$ (fair chance $=0.5$); \textbf{(2) lexical leakage}---a filename token cue scores $0.912$; and \textbf{(3) a difficulty ceiling}---craft domains show near-ceiling human agreement (book/brand/tech $1.0$), and even under A/B randomization (\texttt{eval\_assignment\_seed7}; \texttt{gold\_side\_A} $\approx 0.47$) all five VLM cards reach $1.0$ on craft, with only \textbf{visual} separating ($\approx 0.84$--$0.91$; a post-hoc hard subset, $n=107$, leaves craft at $1.0$). The protocol---expert-only gold, a weak-label firewall, side randomization \emph{at construction}, filename scrubbing, and hard filters---makes each artifact measurable and, for the two construction artifacts, correctable; we release a reference eval harness and a side-randomization tool, and show the tool removes the position skew ($0.916 \rightarrow 0.495/0.505$). The result is a calibrated view of what the benchmark does and does not yet measure: position and leakage are fixable, whereas the craft-domain ceiling is a pair-difficulty limit that Round-2 must address. We do not evaluate engagement, era/circle conditioning, or calibration here, and we claim no general taste SOTA.
\end{abstract}


\section{Introduction}
\label{sec:intro}

Large multimodal models are increasingly asked to \emph{judge}: which poster is better designed, which brand system is more coherent, which book cover fits its brief, which API sketch communicates tradeoffs more clearly. These requests sound like taste, but contemporary training mixtures and reward models often optimize for clicks, dwell, shares, and preference data that themselves correlate with engagement. Collapsing ``what people engage with'' into ``what is good'' risks mistaking outrage, meme velocity, and recommender amplification for craft---the rival \textbf{H3} (engagement-as-taste) in our companion measurement paper.

We take a narrower operational stance. Following the \emph{dubao-taste} program, \textbf{taste} is the ability to discriminate good from bad under an explicit rubric of craft, coherence, and appropriateness to brief. TasteBench asks whether VLMs (and text-only ablations) exhibit that discrimination across domains and under social-style conditioning.

\paragraph{What this paper is.}
We build an expert-gold pairwise craft benchmark and a \textbf{controlled protocol}, quantify three artifacts that undermine its validity, and show which are correctable. The contribution is the \emph{diagnosis and the controls}, not a leaderboard. We scope the release deliberately: era/circle conditioning and calibration are specified but not exercised, and the engagement-as-taste hypothesis is not tested here. Section~\ref{sec:limitations} gives the exact scope of claims.

\paragraph{Contributions.}
\begin{enumerate}[leftmargin=*,itemsep=2pt]
  \item \textbf{A control protocol for trustworthy evaluation.} A small set of controls---expert-only gold, a weak-label firewall, A/B side randomization \emph{at construction}, filename scrubbing, and hard filters on $n\_raters \ge 3$ and \textbf{percent agreement $\ge 0.60$}---each paired with a measurable audit (Section~\ref{sec:control}). The protocol is what lets a taste benchmark's score be read as judgment rather than test-gaming.
  \item \textbf{Three quantified artifacts it exposes.} A measured \emph{position bias} ($P(\text{pref.\ in A})=0.956$; \texttt{always\_item\_a} scores $0.956$), a measured \emph{lexical leak} (\texttt{leak\_prefer\_good\_token} scores $0.912$), and a measured \emph{difficulty ceiling} (craft saturates at $1.0$ for humans and all five VLM cards; only visual separates). The first two are construction artifacts and are correctable; the third is a pair-selection limit.
  \item \textbf{Released contract and tooling.} Schema tables (with \texttt{data\_origin}, \texttt{pair\_group\_id}, \texttt{cue\_type}), \texttt{rubric\_v1}, versioned prompts (\texttt{prompt\_vX}), a reference eval harness (\texttt{preds} CSV $\rightarrow$ metrics JSON), and a side-randomization tool that demonstrably balances the measured subset ($0.916 \rightarrow 0.495/0.505$).
  \item \textbf{A decision-relevant finding.} Under the protocol, Round-1 craft pairs do not yet separate models; we isolate \emph{why} (position, leakage, difficulty) and specify the harder, side-randomized Round-2 required---so the negative result is diagnostic, not terminal.
\end{enumerate}

\paragraph{Non-goals.}
We do \textbf{not} test the engagement-as-taste hypothesis in this paper: no engagement labels were scored, so it remains motivation, not a result. We do not claim that TasteBench measures universal beauty, that expert panels are culture-free, or that high TasteBench scores imply product lift in retrieval or recommendation. A planned Know$\rightarrow$Do follow-on is locked as \textbf{offline Do-slot protocol eval} (plan Option~A in the companion track): TasteBench-aligned VLM scores as Do policy priors vs random / position / engagement-style foils, measured by agreement with the \emph{same} expert gold used here; hard-subset stress as a first cut; \textbf{no InternVL3 weights claim}; live A/B / causal CTR out of scope. \textbf{No product results} are claimed here.

\section{Related Work}
\label{sec:related}

\paragraph{Preference learning and pairwise judgment.}
Bradley--Terry and related pairwise models~\citep{bradley1952rank} underpin modern preference optimization and human-feedback pipelines~\citep{christiano2017deep,ouyang2022instructgpt,rafailov2023dpo}. TasteBench adopts pairwise preference as the primary task because it matches how craft experts often adjudicate close calls; primary pair gold is majority vote among expert raters, with Bradley--Terry reserved for global ranking / analysis / sensitivity when the within-domain pairwise graph is dense enough (mean degree $\ge 3$).

\paragraph{Aesthetic assessment and ``quality'' datasets.}
Computer vision has long predicted aesthetic scores and popularity from photography and design corpora~\citep{murray2012ava,kong2016aadb}. Many such datasets operationalize quality via mean opinion scores that partly track appeal or engagement. TasteBench positions itself against that collapse: engagement may correlate with expert judgment in some slices and diverge sharply in others (controversy, ragebait, SEO packing). TasteBench is designed to treat engagement-trained predictors as baselines and negative controls, not oracle labels; this release does not exercise them (Section~\ref{sec:limitations}).

\paragraph{Cultural capital, habitus, and circle priors.}
Sociological accounts of taste emphasize acquired dispositions and field-specific recognition~\citep{bourdieu1984distinction}. We do not claim a full operationalization of Bourdieu; we only encode \textbf{era} and \textbf{circle} as \emph{conditioning instructions} under which experts and models must judge fit to a stated prior---so that ``good for Y2K poster design'' can differ from ``good for 2020s minimal product UI'' without treating platform traffic as truth.

\paragraph{VLM evaluation beyond recognition.}
Recent multimodal benchmarks such as MMMU, MMBench, and SEED-Bench stress perception, OCR, chart reasoning, and instruction following~\citep{yue2024mmmu,liu2023mmbench,li2023seed}. Fewer isolate \emph{normative craft discrimination} under explicit non-virality rubrics. TasteBench targets that gap with a cross-domain meta-skill framing: the same judgment type (good vs bad under brief) on different surfaces.

\paragraph{Recommender confounding.}
Observational feeds confound preference with position and recommendation~\citep{joachims2017unbiased}. The companion paper studies that identification problem for \emph{people}. TasteBench instead evaluates \emph{models} against expert gold, while allowing engagement signals only as weak train-time material---so success cannot be achieved by reconstructing the engagement head by construction.

\paragraph{Calibration.}
Expected calibration error~\citep{guo2017calibration} is our intended secondary metric; it is specified in the eval contract but \emph{not reported} in this release (Section~\ref{sec:limitations}).

\section{Benchmark Design}
\label{sec:design}

\subsection{Goal and task typology}

TasteBench evaluates whether a model can discriminate \textbf{good vs bad} under \texttt{rubric\_v1}, across domains, optionally conditioned on era/circle. Tasks:

\begin{enumerate}[leftmargin=*,itemsep=2pt]
  \item \textbf{Pairwise preference (primary).} Given items $A$ and $B$ (and optional condition), choose which is better under the rubric. \textbf{Primary pair gold = majority vote} among expert raters. \textbf{Bradley--Terry} is used only when the within-domain pairwise graph has \textbf{mean degree $\ge 3$}, and only for global ranking / analysis / sensitivity tables---it \textbf{does not replace} primary test labels by default.
  \item \textbf{Absolute quality (secondary).} Score an item on $[0,1]$ Likert-calibrated bands; used for calibration analyses and the companion paper's $\tau$ family, not as the primary TasteBench leaderboard number.
  \item \textbf{Conditional taste.} The same content pair under different \texttt{condition\_era} / \texttt{condition\_circle} instructions, linked by \texttt{pair\_group\_id} with \texttt{cue\_type} $\in$ \{\texttt{null}, \texttt{correct}, \texttt{misleading}\}. This measures sensitivity to stated priors---not engagement prediction.
\end{enumerate}

\subsection{Domains (v1)}

\begin{table}[t]
\centering
\small
\begin{tabular}{@{}lll@{}}
\toprule
Domain & Example items & Model inputs \\
\midrule
visual & posters, UI stills, photography & image (+ optional short caption) \\
brand  & logo/identity pairs, packaging & image + brand-agnostic brief \\
book   & covers + blurbs, or passage craft & image and/or text \\
tech   & API snippets, architecture diagrams, product UX & image and/or text \\
\bottomrule
\end{tabular}
\caption{Four Round-1 domains. Cross-domain transfer is intentional: ``can tell good from bad'' is treated as one meta-skill with different surface forms.}
\label{tab:domains}
\end{table}

\subsection{Schema}

Four core tables:

\begin{itemize}[leftmargin=*,itemsep=2pt]
  \item \textbf{\texttt{bench\_items}:} \texttt{item\_id}, \texttt{domain}, \texttt{uri\_or\_payload\_ref}, optional \texttt{era\_tag} / \texttt{circle\_tags}, required \texttt{license} (redistribution-OK assets only), required \texttt{data\_origin}.
  \item \textbf{\texttt{bench\_pairs}:} \texttt{pair\_id}, \texttt{domain}, \texttt{item\_a}, \texttt{item\_b}, optional \texttt{condition\_era} / \texttt{condition\_circle}, optional \texttt{pair\_group\_id} (family for correct vs misleading cue), optional \texttt{cue\_type} $\in$ \{\texttt{null}, \texttt{correct}, \texttt{misleading}\}, \texttt{difficulty} $\in$ \{easy, medium, hard, \textbf{ambiguous}\}, \texttt{split} $\in$ \{train, dev, test\}, optional cached \texttt{agreement} / \texttt{n\_raters}, required \texttt{data\_origin}.
  \item \textbf{\texttt{bench\_gold}:} \texttt{pair\_id}, \texttt{preferred\_id} (pair-grain \textbf{expert majority}; primary test labels), \texttt{agreement} (\textbf{percent agreement}, primary), \texttt{n\_raters}, \texttt{rubric\_id}, \texttt{data\_origin}. The companion paper exposes the same aggregates via \texttt{label\_aggregates} (\texttt{label\_grain} $\in$ \{item, pair\}, \texttt{preferred\_id}, \texttt{exclude\_primary}). Optional BT ranks (within-domain mean degree $\ge 3$) are analysis/sensitivity only---not default \texttt{preferred\_id}.
  \item \textbf{\texttt{bench\_weak} (optional):} \texttt{pair\_id}, \texttt{weak\_preferred\_id}, \texttt{weak\_source}, \texttt{use\_as} $\in$ \{alignment\_only, analysis\_only\}---\textbf{never} test labels; \texttt{data\_origin}.
\end{itemize}

\paragraph{Label policy (frozen).} (1) Test = expert gold only via \texttt{bench\_gold} / \texttt{label\_aggregates}. (2) Weak engagement never enters test metrics (train alignment / analysis only). (3) Simulation stress tests are allowed only with \texttt{data\_origin=simulation} and methods disclosure---never silently merged into primary claims. (4) Eval \textbf{hard-filters} $n\_raters \ge 3$ \emph{and} percent \texttt{agreement} $\ge 0.60$ (inclusive); drops \texttt{difficulty=ambiguous} / \texttt{exclude\_primary} even if tags are wrong.

\subsection{Era and circle conditioning (specified, not exercised in this release)}
\label{sec:conditioning}

Conditions are \emph{instructions}, not hidden features of popularity. Raters and models are told, when a condition is present, to judge fit to that prior; if the condition conflicts with craft, free-text notes are allowed, but the primary label follows rubric $+$ condition. Conditional variants of the same content pair share \texttt{pair\_group\_id}; \texttt{cue\_type} $\in$ \{\texttt{null}, \texttt{correct}, \texttt{misleading}\} marks unconditioned, correctly matched, or deliberately wrong era/circle prompts. Misleading-cue ablations would support the condition-sensitivity metric.

\textbf{Important scope note.} Although the schema materializes 10 \texttt{null}+\texttt{correct}+\texttt{misleading} families (20 conditioned rows), these rows are \texttt{split=dev}, carry \textbf{no expert labels}, appear in \textbf{zero} rows of the primary gold, the eval assignment, or any model prediction, and were never rendered into a cue banner in \texttt{prompt\_v1}. \textbf{Condition sensitivity is therefore a specified design element that is not evaluated in this release.} We report no condition-sensitivity numbers (Section~\ref{sec:limitations}).

\paragraph{Circle tags vs.\ membership.} TasteBench uses item-side \texttt{circle\_tags} (and \texttt{condition\_circle}) as \emph{condition cues}. User circle membership (\texttt{user\_circles}) is the companion paper's primary H2 construct---optional cross-reference only; it is not required to score TasteBench pairs.

\subsection{Relationship to companion work}

The companion measurement paper quantifies \emph{human} taste as the shape of high-quality exposure memory ($\tau$ / half-life / \texttt{user\_circles} H2 live there) and pre-registers H3 (engagement-as-taste) as a rival; both efforts share the expert-gold path (\texttt{label\_aggregates} $\leftrightarrow$ \texttt{bench\_gold}) under the same gold/weak firewall. TasteBench supplies the model-facing expert-gold measurement. A companion Know$\rightarrow$Do track is locked as \textbf{offline Do-slot protocol eval} (plan Option~A); a first $K{=}2$ artifact exists in that track (same expert gold; VLM predictions as Do scores vs random / position / construction foils; H3 engagement $\not\equiv$ quality as negative-control framing; hard-subset stress). Live product A/B / causal CTR remains \textbf{out of scope}---\textbf{this paper cites no companion-track cells as results}.

\section{Annotation Protocol}
\label{sec:annotation}

\subsection{Rubric (\texttt{rubric\_v1})}

Experts answer: given the brief (or domain default), which option is \textbf{better crafted / more coherent / more appropriate}? Explicitly \emph{not} judged: virality, follower count, controversy heat, or platform of origin alone. Domain cues instantiate the same meta-skill:

\begin{itemize}[leftmargin=*,itemsep=2pt]
  \item \textbf{visual:} composition, hierarchy, restraint, technical finish
  \item \textbf{brand:} distinctiveness, consistency, brief-fit (not logo fame)
  \item \textbf{book:} prose craft / cover--content fit / editorial judgment (task-dependent)
  \item \textbf{tech:} clarity of API/UX/architecture tradeoff communication; correctness over hype
\end{itemize}

Absolute anchors $(0$--$1)$: clear fail ($0.0$--$0.3$), competent/generic ($0.4$--$0.6$), strong ($0.7$--$0.85$), exemplary ($0.85$--$1.0$). Pairwise instructions mirror these bands. Guideline edits bump the version id; this release freezes \textbf{\texttt{rubric\_v1}}.

\subsection{Panel process and inclusion rules}

\textbf{Round-1 primary panel.} Raters \texttt{human\_expert\_01} / \texttt{human\_expert\_02} / \texttt{human\_expert\_03} are \textbf{biological human experts}; any agent protocol used for packet delivery is a \textbf{communication channel only} (user-confirmed). Promoted gold path: \texttt{bench\_gold\_primary\_human.csv} (\texttt{gold\_tier=primary\_human\_expert}; all rows \texttt{exclude\_primary=false}). The superseded archival file \texttt{bench\_gold\_provisional\_agent\_panel.csv} is \textbf{not} the current analysis tier.

\begin{enumerate}[leftmargin=*,itemsep=2pt]
  \item At least \textbf{three} independent raters per item/pair; \textbf{no engagement statistics} shown.
  \item Aggregate: \textbf{majority vote = primary pair gold} (\texttt{preferred\_id}); mean for absolute scores. \textbf{Bradley--Terry} only if within-domain pairwise graph \textbf{mean degree $\ge 3$}, and only for global ranking / analysis / sensitivity tables---\textbf{does not replace} primary test labels by default (sparse graphs stay majority-only).
  \item \textbf{Primary agreement gate:} \textbf{percent agreement $\ge 0.60$ (inclusive)} on the discrete label (pairwise \texttt{preferred\_id} or binned absolute). Below gate \emph{or} $n < 3$ $\rightarrow$ \texttt{difficulty=ambiguous} and \texttt{exclude\_primary=true}, \textbf{hard-filtered from the primary test metric} (still retained for analysis of hard cases). \textbf{Secondary (appendix only):} Krippendorff's $\alpha$ (nominal)~\citep{krippendorff2018content}---does not replace the primary gate. Eval must \textbf{hard-filter} on $n\_raters \ge 3$ and percent \texttt{agreement} $\ge 0.60$, not trust difficulty tags alone. (With $n=3$, percent agreement takes only the values $\{1, 2/3, 1/3\}$, so the $\ge 0.60$ gate is equivalent to a $\ge 2/3$ majority.)
  \item Companion defaults ($\tau=0.70$ sweep $\{0.60,0.70,0.80\}$; recency half-life \textbf{90 days}) apply to human exposure / \texttt{memory\_shape} \emph{in the companion paper}---\textbf{cross-ref only}; they are not TasteBench's primary pairwise metric and are not re-frozen here.
\end{enumerate}

\subsection{Weak supervision firewall}

High-interaction platform subsets may yield \texttt{bench\_weak} rows for contrastive warmup or disagreement analysis (\texttt{use\_as=alignment\_only} / \texttt{analysis\_only}). They are barred from test labels and from defining leaderboard success. \textbf{Test gold = expert only} via \texttt{bench\_gold} / \texttt{label\_aggregates}. Engagement-as-taste remains H3: report correlation and divergence (controversy, amplification, outrage), never silently substitute. \emph{In the Round-1 pilot no weak/engagement pairs were evaluated; the firewall is a contract, not an exercised comparison.}

\subsection{Stub corpus (non-empirical)}

The repository ships placeholder CSVs to exercise the contract: 50 items across four domains, 50 pairs (30 test / 10 dev / 10 train), 19 gold pairs meeting $n \ge 3$ and percent agreement $\ge 0.60$, and 10 weak train-only rows. All stub tables carry \texttt{data\_origin} (typically \texttt{simulation} for wiring). Assets are \texttt{CC-BY-4.0-placeholder} paths---\textbf{not a licensed release corpus}. \textbf{No scientific claims are made from stubs; stub/toy numbers are never presented as real results.}

\section{Metrics and Baselines}
\label{sec:metrics}

\subsection{Metrics}

Let $\mathcal{T}$ be the primary test set: gold pairs with \texttt{split=test}, hard-filtered to $n\_raters \ge 3$ \emph{and} \textbf{percent} \texttt{agreement} $\ge 0.60$ (inclusive), dropping \texttt{difficulty=ambiguous} / \texttt{exclude\_primary} (even if tags disagree).

\begin{itemize}[leftmargin=*,itemsep=2pt]
  \item \textbf{Pairwise accuracy (primary).} Fraction of pairs where \texttt{pred\_preferred\_id} equals \texttt{preferred\_id}. Reported overall and by domain.
  \item \textbf{ECE.} If the model emits confidence in $[0,1]$, compute expected calibration error over correctness indicators (default 10 bins)~\citep{guo2017calibration}. \emph{Not reported in this release (no confidence emitted).}
  \item \textbf{Condition sensitivity.} Compare accuracy by \texttt{cue\_type} within each \texttt{pair\_group\_id}---\texttt{correct} vs \texttt{misleading} vs \texttt{null}. \emph{Not reported in this release (no conditioned rows scored).}
  \item \textbf{Cross-domain spread (zero-shot).} Because the Round-1 cards are used zero-shot (no per-domain adaptation), we report the \emph{spread} of per-domain accuracy (max $-$ min) as a consistency diagnostic rather than a transfer matrix (Table~\ref{tab:spread}). A true transfer gap would require domain-adapted runs.
  \item \textbf{Explicit non-metric.} Engagement / click prediction accuracy is \textbf{not} a TasteBench success criterion.
\end{itemize}

\subsection{Baselines}

\begin{enumerate}[leftmargin=*,itemsep=2pt]
  \item \textbf{Strong VLMs} with versioned \texttt{prompt\_vX} templates (pairwise forced choice $+$ optional confidence).
  \item \textbf{Text-ablated} counterparts (captions/briefs only) to separate visual from linguistic priors.
  \item \textbf{Retrieval / aesthetic-score proxies} from prior aesthetic literature~\citep{murray2012ava,kong2016aadb}.
  \item \textbf{Engagement-trained models as negative controls}---systems optimized for likes, saves, or watch time. Competitive engagement prediction with \emph{poor} expert-aligned TasteBench accuracy supports the dissociation thesis; the reverse pattern would favor H3 in that slice and must be reported honestly.
\end{enumerate}

Random / majority-class chance levels are reported per split imbalance. Oracle predictions (copying gold) validate the eval harness only. Empirical non-VLM floors on primary gold ($n=250$) are reported in Section~\ref{sec:results-baselines}; the high \texttt{always\_item\_a} accuracy is a construction artifact, not a ceiling.

\subsection{Eval contract}

\begin{itemize}[leftmargin=*,itemsep=2pt]
  \item \textbf{Input:} predictions CSV with columns \texttt{pair\_id}, \texttt{pred\_preferred\_id}, optional \texttt{confidence}.
  \item \textbf{Gold path:} merge to \texttt{bench\_gold} only (never \texttt{bench\_weak}); filter \texttt{split=test}; \textbf{hard-filter} $n\_raters \ge 3$ \emph{and} percent \texttt{agreement} $\ge 0.60$ (inclusive); drop \texttt{difficulty=ambiguous} / \texttt{exclude\_primary}.
  \item \textbf{Output:} metrics JSON including \texttt{n\_eval}, \texttt{pairwise\_acc}, \texttt{by\_domain}, optional \texttt{ece}, accuracies by \texttt{cue\_type} / \texttt{pair\_group\_id} when present, and a policy string restating the weak-label firewall and \texttt{data\_origin} disclosure.
\end{itemize}

A reference implementation of this contract (reconstructed from the frozen spec; \emph{not} yet reconciled with the original Round-1 runner) ships at \texttt{scripts/tastebench\_eval.py}; it computes pairwise accuracy, by-domain accuracy, ECE (when a \texttt{confidence} column is present), and by-\texttt{cue\_type} slices. A confidence-eliciting template \texttt{prompt\_v2} (and a conditioned variant \texttt{prompt\_cue\_v1}) is provided so ECE and cue sensitivity become computable on the next run, even though they are not reported here.

\section{Protocol}
\label{sec:protocol}

\paragraph{Prompting.} All submissions declare \texttt{prompt\_vX}, model id/revision, decoding settings, and whether images, briefs, or both were provided. Condition strings, when used, must match schema enums or be declared as adversarial/misleading ablations.

\paragraph{Splits.} Train may include weak labels for alignment; dev is for selection; test metrics use expert gold only. Rows with \texttt{data\_origin=simulation}, if any, are disclosed in methods and reported separately---never silently merged into primary claims.

\paragraph{Evaluation / ranking rules (intended).} Primary sort key: pairwise accuracy on $\mathcal{T}$. Secondary: ECE (lower better), mean condition-sensitivity score, and cross-domain spread (smaller is more consistent, with per-domain specialists reported separately). Engagement metrics may appear in an appendix as diagnostics, never as the ranking key.

\paragraph{Reproducibility and release status.} The intended release includes the schema, rubric one-pager, rater guidelines, anonymized gold aggregates, the eval script, the versioned prompts, and the raw prediction files. \textbf{This preprint ships the aggregate tables, per-card scores, two derived diagnostics (side-balance, cross-domain spread), and five figures; it also ships a reference eval harness and the \texttt{prompt\_v2} template.} The raw \texttt{preds\_*.csv} from the Round-1 run and the eval assignment are enumerated in \texttt{RELEASE\_MANIFEST.md} and must accompany the camera-ready for end-to-end reproduction. We flag this rather than overstate reproducibility.

\paragraph{Toy harness check (not a result).} On the stub gold set ($n=19$), a chance-like prediction file yields pairwise accuracy $\approx 0.42$ (toy) with ECE $\approx 0.13$ (toy); an oracle file yields accuracy $1.0$ (harness sanity). These figures validate plumbing only (Appendix~\ref{app:toy}).

\section{A Control Protocol for Expert-Gold Pairwise Evaluation}
\label{sec:control}

The protocol is the paper's primary artifact. It is a small set of controls, each paired with a measurable audit, such that a benchmark's pairwise score can be interpreted as craft judgment rather than test-gaming. None of the controls is novel in isolation; what makes the score trustworthy is the \emph{combination}, together with the requirement that each artifact be audited and reported alongside accuracy.

\begin{table}[t]
\centering
\small
\begin{tabular}{@{}cp{3.3cm}p{2.5cm}p{3.5cm}@{}}
\toprule
\# & Control & Neutralizes & Audit \\
\midrule
C1 & Expert-only gold; weak labels train/analysis-only & virality-as-taste & gold provenance (\texttt{data\_origin}) \\
C2 & Hard filter: $n\_raters \ge 3$ AND percent agreement $\ge 0.60$ & noisy / ambiguous labels & agreement distribution \\
C3 & Randomize A/B side at construction & position bias & side-balance audit (Table~\ref{tab:side}) \\
C4 & Scrub predictive tokens from filenames / IDs & lexical leakage & leak baseline (Table~\ref{tab:baselines}) \\
C5 & Report difficulty; reject or replace saturated domains & difficulty ceiling & per-domain spread (Table~\ref{tab:spread}) \\
\bottomrule
\end{tabular}
\caption{The control protocol. Each control is paired with an audit that makes the artifact visible before the score is trusted. C1--C2 are enforced by the released eval harness; C3 is implemented by \texttt{randomize\_pair\_sides.py}; C4 is a data-hygiene step; C5 is a reporting requirement.}
\label{tab:tcp}
\end{table}

\textbf{How to read a TasteBench score.} A score is meaningful only together with its audits: report the side-balance audit (C3), the leak-baseline score (C4), and the per-domain spread (C5) next to pairwise accuracy. In our Round-1 data, C3 and C4 show two trivial policies scoring $0.956$ and $0.912$ (Section~\ref{sec:results-baselines})---without them, almost any leaderboard number is uninterpretable.

\section{Results}
\label{sec:results}

\subsection{Primary panel and gold (descriptive)}
\label{sec:results-panel}

\textbf{Disclosure (required).} Round-1 primary gold path: \path{labeling/round1/aggregates/bench_gold_primary_human.csv} (tier \texttt{primary\_human\_expert}). Rater IDs \texttt{human\_expert\_01}/\texttt{02}/\texttt{03} are \textbf{biological human experts}; the agent protocol is a \textbf{communication channel only}. Every primary gold row is tagged \texttt{gold\_tier=primary\_human\_expert}, \texttt{data\_origin=study\_human\_expert}, and \texttt{exclude\_primary=false}. The older provisional agent-panel file is \textbf{superseded archival only}.

\begin{table}[t]
\centering
\small
\begin{tabular}{@{}lr@{}}
\toprule
Quantity & Value \\
\midrule
Label rows & 750 \\
Gold pairs (primary) & 250 \\
Raters & \texttt{human\_expert\_01}, \texttt{02}, \texttt{03} ($n$\_raters $=3$ on all pairs) \\
Primary-eligible & 250 / 250 \\
Ambiguous & 0 \\
Unanimous triples & 247 / 250 \\
Rubric & \texttt{rubric\_v1} \\
Intent QC match & $1.0$ on 225 pairs (construction check only) \\
\bottomrule
\end{tabular}
\caption{Primary panel / gold coverage. Intent QC is a construction check, not gold and not a model score.}
\label{tab:panel}
\end{table}

\begin{table}[t]
\centering
\small
\begin{tabular}{@{}lrr@{}}
\toprule
Domain & Pairs & Mean agreement \\
\midrule
book   & 60 & $1.0$ \\
brand  & 60 & $1.0$ \\
tech   & 50 & $1.0$ \\
visual & 80 & $\approx 0.9875$ \\
\bottomrule
\end{tabular}
\caption{Domain mean agreement. Craft domains are unanimous; only three visual pairs are not.}
\label{tab:domain-agree}
\end{table}

\begin{table}[t]
\centering
\small
\begin{tabular}{@{}lr@{}}
\toprule
Agreement & Pairs \\
\midrule
$1.0$    & 247 \\
$0.6667$ & 3 \\
\bottomrule
\end{tabular}
\caption{Agreement distribution. The three non-unanimous pairs are all Met \textbf{visual} (\texttt{pair\_visual\_met\_013}, \texttt{\_015}, \texttt{\_016}).}
\label{tab:agree-dist}
\end{table}

\begin{figure}[t]
\centering
\includegraphics[width=0.60\linewidth]{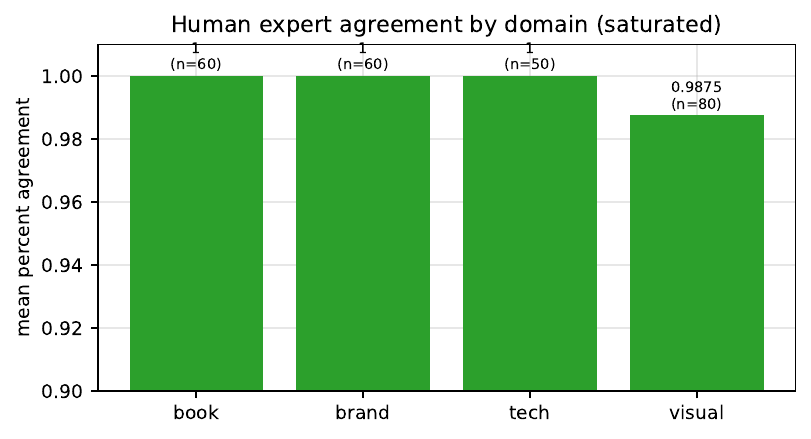}
\caption{Human expert mean percent agreement by domain. Craft domains are saturated at $1.0$; the Round-1 contrasts are easy.}
\label{fig:agreement}
\end{figure}

Primary gold is \textbf{ready for controlled evaluation} under the expert-only test policy. Near-ceiling agreement may still reflect easy good/bad craft contrasts; it must not be cited as a VLM performance ceiling. Intent QC $\ne$ gold.

\subsection{Non-VLM baselines (primary gold; $n = 250$)}
\label{sec:results-baselines}

These are \textbf{non-VLM} floors / diagnostics only---\textbf{not} a model leaderboard and \textbf{not} VLM numbers.

\begin{table}[t]
\centering
\small
\begin{tabular}{@{}lrrrrrr@{}}
\toprule
baseline & $n$ & acc & book & brand & tech & visual \\
\midrule
\texttt{always\_item\_a}        & 250 & 0.956 & 1.0 & 1.0 & 1.0 & 0.8625 \\
\texttt{always\_item\_b}        & 250 & 0.044 & 0.0 & 0.0 & 0.0 & 0.1375 \\
\texttt{random\_seed42}         & 250 & 0.552 & 0.5833 & 0.55 & 0.6 & 0.5 \\
\texttt{hash\_coin}             & 250 & 0.500 & 0.5667 & 0.45 & 0.44 & 0.525 \\
\texttt{leak\_prefer\_good\_token} & 250 & 0.912 & 0.8333 & 1.0 & 1.0 & 0.85 \\
\bottomrule
\end{tabular}
\caption{Non-VLM floors on primary gold. \texttt{always\_item\_a} high accuracy is a construction artifact, not a ceiling; fair floors are \texttt{random\_seed42} / \texttt{hash\_coin}; \texttt{leak\_prefer\_good\_token} is a leakage diagnostic only.}
\label{tab:baselines}
\end{table}

\begin{figure}[t]
\centering
\includegraphics[width=0.64\linewidth]{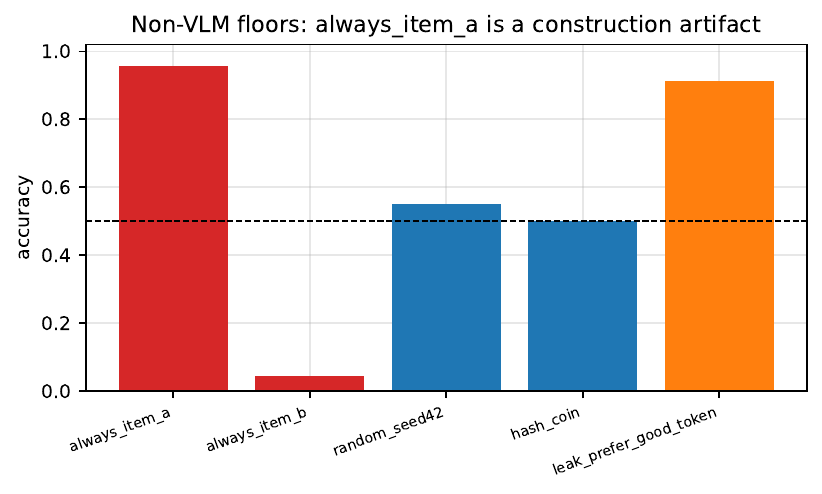}
\caption{Non-VLM floors on primary gold. \texttt{always\_item\_a} and \texttt{leak\_prefer\_good\_token} are artifacts; fair floors are \texttt{random\_seed42} / \texttt{hash\_coin}.}
\label{fig:baselines}
\end{figure}

\textbf{Required disclosure.} The high accuracy of \texttt{always\_item\_a} (0.956 overall; 1.0 on book/brand/tech) is a \textbf{dataset construction artifact}: preferred / ``good'' items are often placed as \texttt{item\_a} in the Round-1 pair files. It is \textbf{not} a model performance ceiling and must not be cited as one. Symmetrically, \texttt{always\_item\_b} (0.044) is the complementary artifact.

\textbf{Fair floors.} Treat \texttt{random\_seed42} (0.552) and \texttt{hash\_coin} (0.5) as the fair chance / non-informative floors.

\textbf{Filename-cue diagnostic only.} \texttt{leak\_prefer\_good\_token} (0.912) exploits a self-authored filename token cue; it is \textbf{not} a fair baseline---report only as a leakage diagnostic.

\textbf{Recommendation.} Future annotation / construction rounds should \textbf{randomize A/B side} (preferred as \texttt{item\_a} vs \texttt{item\_b}) so side-biased trivial policies cease to look like ceilings. Tooling is provided (\texttt{scripts/randomize\_pair\_sides.py}).

\textbf{Side-balance audit and the effect of randomization.} Table~\ref{tab:side} makes the position artifact quantitative. On the full 250-pair gold the preferred item sits in slot A with probability $0.956$ (book/brand/tech $1.0$; visual $0.8625$). The 107-pair hard subset, whose per-pair slots are available, shows the same skew ($0.916$ overall). Applying the released side-randomization tool to that subset (seed 42) balances the sides to $0.4953/0.5047$, turning the position-only floor from a strong-looking baseline into unbiased chance. Re-deriving the full 250 requires the Round-1 pair file; the protocol is to randomize side \emph{at construction} in Round-2.

\begin{table}[t]
\centering
\small
\begin{tabular}{@{}llrrr@{}}
\toprule
scope & domain & $n$ & P(pref.\ is A) & P(pref.\ is B) \\
\midrule
full 250 & all              & 250 & 0.9560 & 0.0440 \\
full 250 & book/brand/tech  & 170 & 1.0    & 0.0    \\
full 250 & visual           & 80  & 0.8625 & 0.1375 \\
hard 107 & all              & 107 & 0.9159 & 0.0841 \\
hard 107 & visual           & 40  & 0.775  & 0.225  \\
\midrule
hard 107 (side-randomized, seed 42) & all & 107 & 0.4953 & 0.5047 \\
\bottomrule
\end{tabular}
\caption{Side-balance audit. The preferred item sits in slot A far more often than chance; the released randomization tool (\texttt{randomize\_pair\_sides.py}) restores balance on the subset with per-pair slots. Sources: \texttt{tastebench\_baselines\_summary.csv}, \texttt{craft\_hard\_subset.csv}.}
\label{tab:side}
\end{table}

\subsection{Round-1 VLM results (the ceiling in practice)}
\label{sec:results-vlm}

\textbf{Protocol.} Provider = OpenRouter chat completions; assignment = \texttt{eval\_assignment\_seed7.csv} ($n=250$, seed $=7$; \textbf{gold\_side\_A rate $\approx 0.47$}---A/B randomized so the construction artifact of Section~\ref{sec:results-baselines} is mitigated for VLM scoring); prompt = \texttt{prompt\_v1}; model map locked at card version v1.2. Pairwise accuracy overall and by domain against \texttt{bench\_gold\_primary\_human.csv}.

\begin{table}[t]
\centering
\small
\begin{tabular}{@{}llrrrrr@{}}
\toprule
model card & runtime \texttt{openrouter\_id} & acc & book & brand & tech & visual \\
\midrule
gpt-5.6-terra     & \texttt{openai/gpt-5.6-terra}      & 0.960 & 1.0 & 1.0 & 1.0 & 0.875 \\
claude-sonnet-5   & \texttt{anthropic/claude-sonnet-5} & 0.956 & 1.0 & 1.0 & 1.0 & 0.8625 \\
gemini-3.7-flash  & \texttt{google/gemini-3.7-flash}   & 0.972 & 1.0 & 1.0 & 1.0 & 0.9125 \\
Qwen3-VL-8B       & \texttt{qwen/qwen3-vl-8b-instruct} & 0.948 & 1.0 & 1.0 & 1.0 & 0.8375 \\
InternVL3-8B (card) & \texttt{google/gemma-3-12b-it}   & 0.968 & 1.0 & 1.0 & 1.0 & 0.900 \\
\bottomrule
\end{tabular}
\caption{Round-1 VLM results. Craft domains are at ceiling for all cards; only visual separates. InternVL3-8B is a disclosed runtime substitute.}
\label{tab:vlm}
\end{table}

\begin{figure}[t]
\centering
\includegraphics[width=0.92\linewidth]{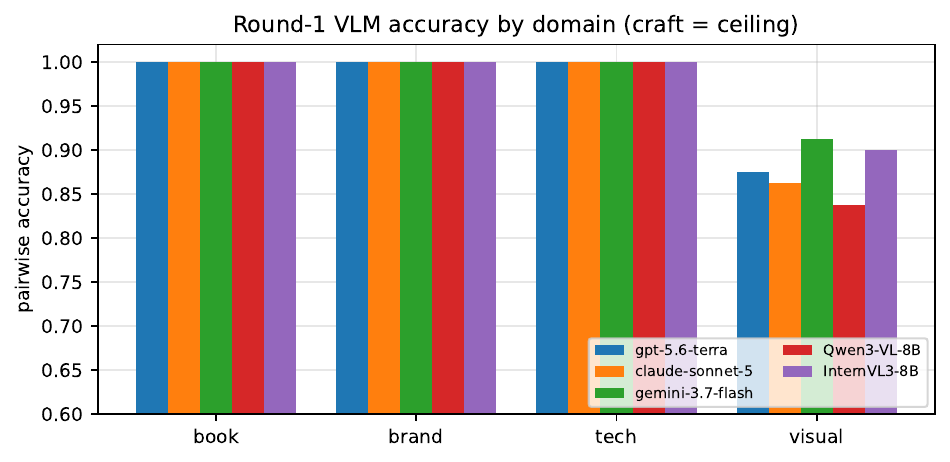}
\caption{Round-1 VLM accuracy by domain. Craft domains are at ceiling for every card; only visual separates.}
\label{fig:acc-domain}
\end{figure}

\textbf{Required disclosure --- InternVL3-8B runtime substitute.} The locked map intended \texttt{nvidia/nemotron-nano-12b-v2-vl:free} for the InternVL3-8B card; the free tier stalled / rate-limited ($\approx 12/250$ in $\approx 3$ min), so the finished predictions for that card are from \texttt{google/gemma-3-12b-it}. Mark as \textbf{runtime substitute}. Do \textbf{not} claim InternVL3 weights were evaluated. Four other cards ran on their locked OpenRouter IDs.

\textbf{Limitations of this Round-1 slice.} Book / brand / tech $= 1.0$ for all five cards likely reflects easy craft contrasts plus near-ceiling human IRR (Section~\ref{sec:results-panel}), not universal mastery. \textbf{Visual} is where models separate ($\approx 0.84$--$0.91$). Compare to fair chance \texttt{hash\_coin} $= 0.5$. Do \textbf{not} claim general SOTA taste without harder pairs, contested items, or broader domain coverage. ECE and condition sensitivity are \textbf{not reported} for this run (no confidence column $\rightarrow$ ECE not computable; cue triples are $0$ in gold/preds $\rightarrow$ cue $\Delta$acc not scorable); zero-shot cross-domain spread is reported in Table~\ref{tab:spread}. \textbf{All accuracies are point estimates and we do not significance-test model differences:} at $n=250$ ($n=50$--$80$ per domain) the 95\% binomial half-width is roughly $\pm 0.03$ ($\pm 0.08$ per domain), so small inter-card gaps (e.g.\ gpt-5.6-terra and claude-sonnet-5 differ by a single pair) are within noise.

\textbf{Post-hoc craft hard subset (stress slice; not new gold).} Composition: \textbf{visual 40 / book 27 / brand 20 / tech 20 $= 107$} pairs; every row keeps primary-gold \texttt{preferred\_id} with \texttt{include\_in\_eval=true}. Primary gold is unchanged. \textbf{Honest shortfall:} because Round-1 craft VLMs sit at ceiling on book/brand/tech (accuracy $=1.0$), the brand/tech hard slices are \textbf{low-expert-confidence only} (weak discrimination; padded to $n=20$); book uses filename-cue fails $+$ low confidence---\textbf{not} VLM-error hardness. Scored post-hoc from existing predictions; because the visual slice was selected using the same models' errors, its numbers are descriptive rather than an independent estimate of skill:

\begin{table}[t]
\centering
\small
\begin{tabular}{@{}lrrrrrr@{}}
\toprule
model card & hard $n$ & hard acc & visual (40) & book (27) & brand (20) & tech (20) \\
\midrule
gemini-3.7-flash          & 107 & 0.9346 & 0.825 & 1.0 & 1.0 & 1.0 \\
InternVL3-8B (card/gemma) & 107 & 0.9252 & 0.800 & 1.0 & 1.0 & 1.0 \\
gpt-5.6-terra             & 107 & 0.9065 & 0.750 & 1.0 & 1.0 & 1.0 \\
claude-sonnet-5           & 107 & 0.8972 & 0.725 & 1.0 & 1.0 & 1.0 \\
Qwen3-VL-8B               & 107 & 0.8785 & 0.675 & 1.0 & 1.0 & 1.0 \\
\bottomrule
\end{tabular}
\caption{Post-hoc hard stress subset. Craft domains remain at $1.0$, consistent with the ceiling shortfall. Cite as stress disclosure only; do not replace full primary-gold metrics.}
\label{tab:hardsubset}
\end{table}

\begin{figure}[t]
\centering
\includegraphics[width=0.92\linewidth]{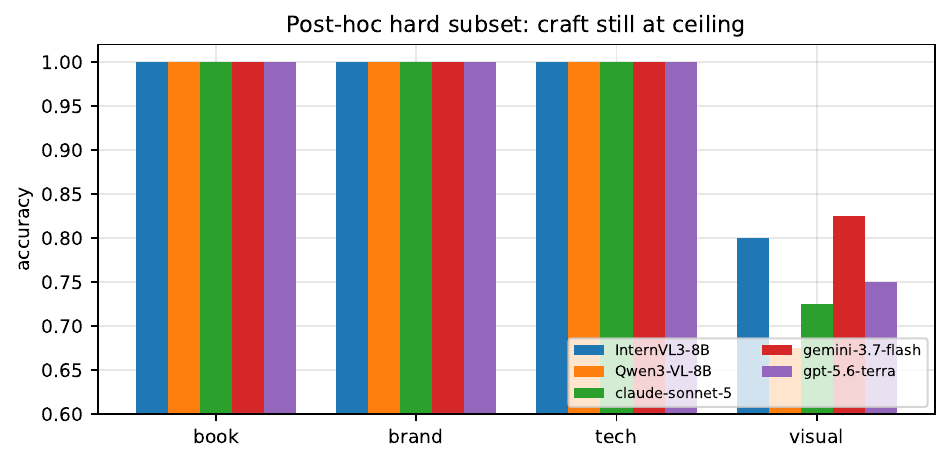}
\caption{Post-hoc hard subset: craft domains still at $1.0$; only visual moves.}
\label{fig:hardsubset}
\end{figure}

\textbf{Zero-shot cross-domain spread.} Because all cards are used zero-shot, the honest generalization number is the spread of per-domain accuracy (max $-$ min), not a transfer matrix:

\begin{table}[t]
\centering
\small
\begin{tabular}{@{}lrrrrr@{}}
\toprule
model card & book & brand & tech & visual & spread \\
\midrule
gemini-3.7-flash       & 1.0 & 1.0 & 1.0 & 0.9125 & 0.0875 \\
InternVL3-8B (gemma)   & 1.0 & 1.0 & 1.0 & 0.900  & 0.1000 \\
gpt-5.6-terra          & 1.0 & 1.0 & 1.0 & 0.875  & 0.1250 \\
claude-sonnet-5        & 1.0 & 1.0 & 1.0 & 0.8625 & 0.1375 \\
Qwen3-VL-8B            & 1.0 & 1.0 & 1.0 & 0.8375 & 0.1625 \\
\bottomrule
\end{tabular}
\caption{Zero-shot cross-domain spread (max domain accuracy $-$ min). Sources: \texttt{vlm\_main\_table.csv}, \texttt{cross\_domain\_spread.csv}.}
\label{tab:spread}
\end{table}

\begin{figure}[t]
\centering
\includegraphics[width=0.60\linewidth]{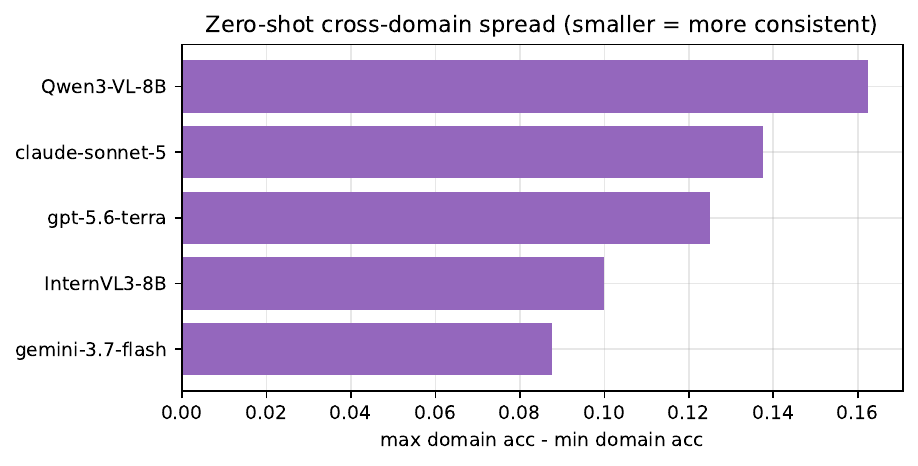}
\caption{Zero-shot cross-domain spread per card (smaller is more consistent).}
\label{fig:spread}
\end{figure}

\textbf{Roster context.} Explicitly not in the main table: GPT-4o / 4o-mini; Claude 3.5--4.x as primary; Gemini 1.5 / 2.5 as primary; Qwen2 / Qwen2.5-VL; classic LLaVA-OneVision-7B. Stub/toy harness checks (Appendix~\ref{app:toy}) remain \textbf{plumbing validation only} and are \textbf{not} leaderboard entries.

\section{Beyond TasteBench: recurring artifacts and a checklist}
\label{sec:beyond}

The three artifacts are not specific to our pairs. Any pairwise ``good-vs-bad'' benchmark assembled from curated assets with semantic identifiers and an expert panel can face them, so the audits generalize:

\begin{table}[t]
\centering
\small
\begin{tabular}{@{}lp{4.4cm}p{4.2cm}@{}}
\toprule
Artifact & Mechanism & Early detection \\
\midrule
Position bias & A/B slot prior in construction or in the judging interface & report $P(\text{pref.\ in A})$; it should be $\approx 0.5$ \\
Lexical leakage & Predictive tokens in filenames / IDs (\texttt{good\_}, \texttt{real\_}, \texttt{v2\_final}) & train a filename-only baseline; it should be $\approx$ chance \\
Difficulty ceiling & Consensus pairs an expert panel cannot make hard & report per-domain accuracy; a saturated domain is \emph{unmeasured}, not solved \\
\bottomrule
\end{tabular}
\caption{Recurring artifacts in pairwise expert benchmarks, independent of domain or model.}
\label{tab:recurring}
\end{table}

\textbf{Why only visual separates.} The craft domains saturate because the Round-1 pairs are easy (Section~\ref{sec:results-panel})---\emph{not} because of position: the VLM eval used a side-balanced assignment (\texttt{gold\_side\_A} $\approx 0.47$), under which a position-only policy scores $\approx 0.47$, far below the observed $1.0$. Position bias affects the \emph{trivial-baseline} layer of the original construction; the craft saturation is a difficulty limit. Visual is the only domain whose pairs carry a genuine accuracy gradient.

\textbf{Checklist for building an expert-gold pairwise benchmark.}
\begin{quote}
\begin{enumerate}[leftmargin=*,itemsep=2pt]
  \item \textbf{Randomize A/B side at construction} and report the side-balance audit; a position-only policy must score $\approx$ chance.
  \item \textbf{Scrub predictive tokens} from filenames, IDs, and paths; a filename-only baseline must score $\approx$ chance.
  \item \textbf{Freeze expert-only gold} and keep any engagement signal train/analysis-only.
  \item \textbf{Hard-filter} on $n\_raters \ge 3$ \emph{and} percent agreement $\ge 0.60$; report the agreement distribution.
  \item \textbf{Report per-domain accuracy}, and treat a saturated domain (all models at ceiling) as \emph{unmeasured}, not as solved.
  \item \textbf{Report the trivial-policy ceiling} (\texttt{always\_item\_a}, filename leak) next to model scores.
  \item \textbf{Release the audits and the harness}, not only the accuracy table.
\end{enumerate}
\end{quote}

\textbf{Applicability to platform-scale and social-media data.} The controls are agnostic to where a corpus comes from, so they transfer directly to platform-scale and social-media data. A content platform is in some ways \emph{better} positioned than this academic pilot: it can assemble large expert-gold sets, and it owns the engagement logs needed to run the negative control we could not (Section~\ref{sec:limitations})---mapping where engagement and expert quality diverge. The discipline is unchanged: engagement may inform analysis but must not enter the gold, sides must be randomized at construction, and identifiers must be scrubbed before release. Applied this way, the protocol turns a private corpus into a trustworthy craft-judgment benchmark; applied with engagement as the answer key, it reproduces exactly the artifacts above.

\section{Analysis Plan}
\label{sec:analysis}

Pre-registered analyses (aligned with frozen defaults):

\begin{enumerate}[leftmargin=*,itemsep=2pt]
  \item \textbf{Discrimination vs engagement.} Correlate model preference with expert gold vs with weak engagement labels on the same pairs; report regions of agreement and conflict. \emph{(Not exercised in Round-1.)}
  \item \textbf{Difficulty slices.} Accuracy on easy/medium/hard gold pairs; separate reporting for ambiguous (excluded) pairs. The post-hoc craft hard subset ($n=107$) is a disclosed stress slice---\textbf{not} new gold.
  \item \textbf{Prior sensitivity.} Within \texttt{pair\_group\_id}, does \texttt{cue\_type=correct} improve expert-aligned accuracy vs \texttt{misleading}/\texttt{null} more for VLMs than for engagement-trained controls? \emph{(Not scorable in Round-1: no conditioned rows scored.)}
  \item \textbf{Modality ablations.} Image-only vs text-only vs both, by domain.
  \item \textbf{Error taxonomy.} Manual codes: brief mismatch, era anachronism, fame/brand recognition bias, toxicity/virality bias, technical incorrectness.
  \item \textbf{Robustness.} Prompt paraphrase sets under the same \texttt{prompt\_vX} family; seed variance for sampling decoders.
  \item \textbf{Human ceiling.} Panel self-consistency and held-out expert agreement as soft upper bounds---not proof of objectivity.
\end{enumerate}

\section{Limitations}
\label{sec:limitations}

\begin{itemize}[leftmargin=*,itemsep=3pt]
  \item \textbf{Expert panels are situated.} Rubric-based craft judgment still reflects who is recruited; era/circle tags cannot exhaust cultural space.
  \item \textbf{Agreement threshold.} Excluding percent agreement $< 0.60$ improves metric stability but may discard theoretically interesting contested taste; $\alpha$ remains appendix-only.
  \item \textbf{Domain coverage.} Four domains are illustrative, not exhaustive (music, fashion film, cuisine, etc., deferred).
  \item \textbf{Conditioning not evaluated.} Era/circle prompts (\texttt{cue\_type} on \texttt{circle\_tags} / era cues) are specified but \textbf{not exercised} in this release; no condition-sensitivity number is reported. No causal claim about lived subcultural membership is made.
  \item \textbf{Weak labels not exercised.} Platform engagement filters are noisy and platform-specific; they exist only for alignment/analysis, and none were evaluated in Round-1.
  \item \textbf{No product causal claim.} Mapping TasteBench competence $\rightarrow$ Know$\rightarrow$Do is offline Do-slot protocol eval (plan Option~A) in the companion track---same expert gold; VLM predictions as policy scores; engagement-style foils as negative controls; live product A/B out of scope. \textbf{Not demonstrated in this paper's Results; no companion-track cells cited here.}
  \item \textbf{Stub assets.} Current public stubs are placeholders; redistribution requires cleared licenses.
  \item \textbf{Craft-domain ceiling.} Round-1 VLM \textbf{book/brand/tech accuracy $= 1.0$} (all five cards) alongside human domain agreement $= 1.0$ (visual human $\approx 0.9875$; only three Met-visual disagreements at $0.6667$) is best read as an \textbf{easy-pair ceiling}---\textbf{not} general taste SOTA or universal mastery. Visual is where models separate ($\approx 0.84$--$0.91$).
  \item \textbf{Post-hoc hard subset is a stress slice, not new gold.} Craft domains have \textbf{no VLM errors}, so brand/tech hard slices are low-confidence only (honest shortfall); visual includes Met soft-gold $013/015/016$ $+$ VLM-error ranking. Hard-subset overall accuracy $\approx 0.88$--$0.93$ (visual $\approx 0.675$--$0.825$; book/brand/tech still $1.0$). Cite as disclosure, not as replacement leaderboard.
  \item \textbf{Met soft-gold / no fourth adjudicator.} Met visual $013/015/016$ are $2/3$ majority soft-gold; adjudication CSVs document votes; \textbf{no fourth rater}; primary gold keeps majority preference.
  \item \textbf{Position/construction artifact.} Preferred/good items were often placed as \texttt{item\_a}, so \texttt{always\_item\_a} $\approx 0.956$ is a \textbf{construction artifact}, not a model ceiling; fair floors are \texttt{random\_seed42} $\approx 0.552$ / \texttt{hash\_coin} $= 0.5$. The VLM run mitigates this via \texttt{eval\_assignment\_seed7} A/B randomization (\texttt{gold\_side\_A} $\approx 0.47$); future rounds should randomize side at construction.
  \item \textbf{InternVL3-8B runtime substitute.} The card is \texttt{google/gemma-3-12b-it} after a free-tier stall---\textbf{not} InternVL3 weights. ECE and condition sensitivity are \textbf{not reported} for this run (predictions lack confidence; cue schema has triples but $0$ conditioned rows in gold / assignment / predictions), though the shipped eval harness and \texttt{prompt\_v2} / \texttt{prompt\_cue\_v1} now make both computable on a re-run; zero-shot cross-domain spread is reported (Table~\ref{tab:spread}). Do \textbf{not} invent ECE or cue-sensitivity numbers from this run.
  \item \textbf{Companion observational proxy.} The companion paper's observational $q$ is a percentile rank of random \texttt{long\_view\_rate}, not expert aesthetic gold---do not import it as TasteBench success.
\end{itemize}

\section{Ethics and Disclosure}
\label{sec:ethics}

\begin{itemize}[leftmargin=*,itemsep=2pt]
  \item Use \textbf{licensed} or otherwise redistribution-permitted assets; record \texttt{license} per item.
  \item Do not show raters engagement statistics during gold labeling.
  \item Hash or anonymize rater identifiers in public releases; obtain consent consistent with institutional policy.
  \item Disclose any \texttt{data\_origin=simulation} rows used for stress tests; never present simulation as human gold.
  \item Disclose Round-1 gold tier explicitly: \texttt{primary\_human\_expert} raters are biological humans (agent protocol = communication channel only). Do not narrate the superseded archival agent-panel file as current analysis gold.
  \item Dual-use: craft-discrimination tools can be abused for persuasive design or engagement hacking; we explicitly discourage optimizing TasteBench via engagement proxies and document negative controls to surface that failure mode.
  \item Cultural respect: circle tags are analytical labels, not pejoratives; guidelines forbid rewarding stereotypes as ``taste.''
\end{itemize}

\section{Conclusion}
\label{sec:conclusion}

Expert-aligned taste benchmarks should reward craft judgment, not test-gaming. TasteBench pairs expert-only gold with a controlled protocol---a weak-label firewall, side randomization at construction, filename scrubbing, and hard filters---that makes the difference measurable. Applied to a Round-1 pilot (\texttt{primary\_human\_expert}; three biological human experts; $n=250$), the protocol quantifies three artifacts---position bias (\texttt{always\_item\_a} $\approx 0.956$), lexical leakage (\texttt{leak\_prefer\_good\_token} $\approx 0.912$), and a difficulty ceiling (craft saturated at $1.0$ for humans and all five VLM cards; only visual separates, $\approx 0.84$--$0.91$)---and shows that the two construction artifacts are \emph{correctable} while the ceiling is a pair-difficulty limit. We release the eval harness, the side-randomization tool, and the prompt templates. The contribution is this controlled view: it tells practitioners which pitfalls to remove, and what must change---harder, contested, side-randomized pairs---before such a benchmark can support general taste claims. We deliberately decline the claims we cannot yet support (engagement, era/circle conditioning, calibration, general SOTA).


\appendix

\section{Frozen defaults (cross-paper)}
\label{app:defaults}

\textbf{TasteBench (this paper):} expert $n \ge 3$; \textbf{percent agreement $\ge 0.60$ (inclusive)} primary gate; below $\rightarrow$ \texttt{difficulty=ambiguous} / \texttt{exclude\_primary}; Krippendorff $\alpha$ appendix-only; hard-filter eval; weak $\rightarrow$ train/alignment only; test = expert (\texttt{bench\_gold} / \texttt{label\_aggregates} with \texttt{label\_grain} / \texttt{preferred\_id}); all bench tables carry \texttt{data\_origin}; conditional pairs use \texttt{pair\_group\_id} $+$ \texttt{cue\_type} $\in$ \{\texttt{null}, \texttt{correct}, \texttt{misleading}\}; H3 engagement-as-taste is a competing hypothesis, not gold. \textbf{Companion-only (cross-ref, not re-frozen here):} $\tau=0.70$ (sweep $0.60/0.80$); recency half-life $90$d (sweep $30/90/180$); H2 circle membership $=$ \texttt{user\_circles} (TasteBench may use item \texttt{circle\_tags} as condition cues only).

\section{Toy eval (non-empirical)}
\label{app:toy}

We validated the eval harness on a 19-pair stub gold set: a chance-like predictor yields pairwise accuracy $\approx 0.42$ and ECE $\approx 0.13$, while an oracle predictor yields $1.0$. These figures confirm that the pipeline computes its metrics correctly; \textbf{they are not model performance}, and the stub set is a wiring fixture, not empirical data.

\bibliographystyle{plainnat}
\bibliography{refs}

\end{document}